\pdfoutput=1
\documentclass[11pt]{article}

\usepackage[]{ACL2023}

\usepackage{times}
\usepackage{latexsym}

\usepackage[T1]{fontenc}
\usepackage[utf8]{inputenc}

\usepackage{microtype}

\usepackage{inconsolata}

\usepackage{graphicx}
\usepackage{amsmath}
\usepackage{amssymb}
\usepackage{booktabs}
\usepackage{bm}
\usepackage{multirow}
\usepackage{color}
\title{Multijugate Dual Learning for Low-Resource \\Task-Oriented Dialogue System}

\author{
    Shimin Li\textsuperscript{\rm 1},
    Xiaotian Zhang\textsuperscript{\rm 1},
    Yanjun Zheng\textsuperscript{\rm 1},
    Linyang Li\textsuperscript{\rm 1},
    Xipeng Qiu\textsuperscript{\rm 1,2}\thanks{Corresponding Author.}, \\
    \textsuperscript{\rm 1} School of Computer Science, Fudan University\\
    \textsuperscript{\rm 2} Shanghai Key Laboratory of Intelligent Information Processing, Fudan University\\
    \texttt{smli20@fudan.edu.cn, \{xiaotianzhang21, yanjunzheng21\}@m.fudan.edu.cn}, \\
    \texttt{\{linyangli19, xpqiu\}@fudan.edu.cn} \\
}

\begin{document}
\maketitle
\begin{abstract}

Dialogue data in real scenarios tend to be sparsely available, rendering data-starved end-to-end dialogue systems trained inadequately. We discover that data utilization efficiency in low-resource scenarios can be enhanced by mining alignment information uncertain utterance and deterministic dialogue state. Therefore, we innovatively implement dual learning in task-oriented dialogues to exploit the correlation of heterogeneous data. In addition, the one-to-one duality is converted into a multijugate duality to reduce the influence of spurious correlations in dual training for generalization. Without introducing additional parameters, our method could be implemented in arbitrary networks. Extensive empirical analyses demonstrate that our proposed method improves the effectiveness of end-to-end task-oriented dialogue systems under multiple benchmarks and obtains state-of-the-art results in low-resource scenarios.

\end{abstract}

\section{Introduction}
With the emergence of dialogue data \cite{dialogpt}, and the evolution of pre-trained language models \cite{survey_pretrain}, end-to-end task-oriented dialogue (TOD) systems \cite{pptod,mttod,q-tod} gradually replaced the previous modular cascading dialogue systems \cite{survey_conv_ai}. The end-to-end TOD system adopts a uniform training objective, preventing the error propagation problem in pipelined dialogue systems \cite{survey_conv_ai}. Nonetheless, the end-to-end paradigm requires more training data to perform better \cite{pptod}.
Meanwhile, TOD data is enormously expensive to annotate \cite{multiwoz} as it simultaneously contains dialogue state tracking, dialogue action prediction, and response generation. It is also expensive to annotate large amounts of complicated dialogue data for each emerging domain \cite{cins}. Therefore, improving data utilization efficiency in low-resource scenarios becomes critical for end-to-end TOD.

\begin{figure}[t]
\centering
\includegraphics[width=1\columnwidth]{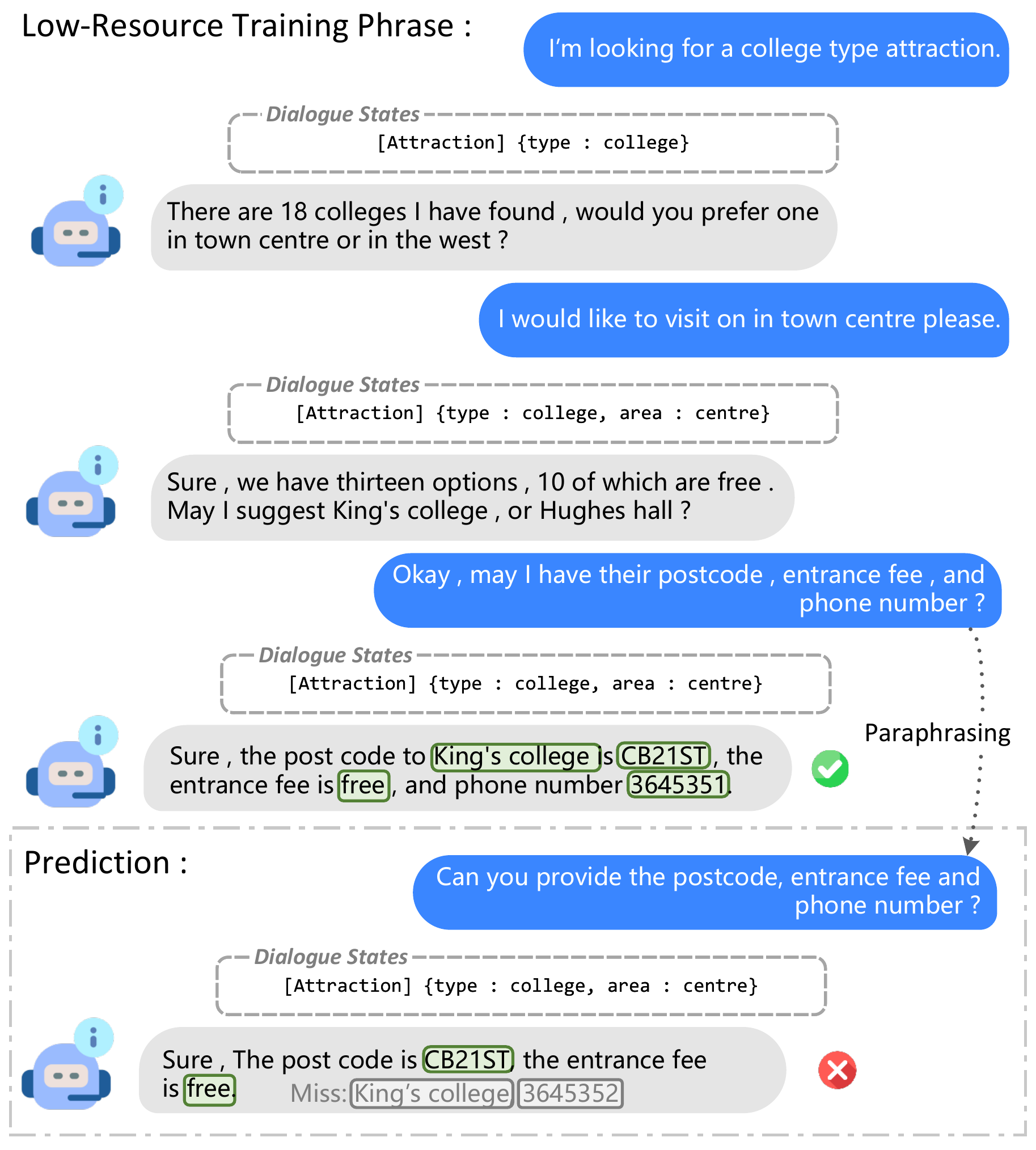} % Reduce the figure size so that it is slightly narrower than the column. Don't use precise values for figure width.This setup will avoid overfull boxes.
\caption{The TOD training and prediction procedure in the low-resource scenario. When the user utterance is rephrased, the predictions miss some entities.}
\label{fig:task_intro}
\end{figure}

Previous approaches \cite{dialogpt,pptod} improve the transferability of models on downstream tasks and capacity to handle small samples by conducting self-supervised or semi-supervised further-pretraining \cite{galaxy} of models on data from additional dialogue domains. However, these further pre-trains on million-level datasets may require hundreds of GPU hours and are resource-constrained.
Then on specific downstream dialogue tasks, a unified multi-task generative paradigm \cite{mttod,pptod} was applied to end-to-end dialogue tasks. Although this generative approach demonstrates better generalization and outcomes, we argue that heterogeneity and duality between data are ignored. Here, heterogeneity refers to the formative discrepancy between uncertain, unstructured discourse (e.g., user utterances and system responses) and deterministic, structured dialogue states.
Accordingly, the underlying alignment information and knowledge contained within the heterogeneous data is not fully exploited in the above approach.

% To tackle the above challenges, we propose an innovative multijugate dual learning framework in TOD (MDTOD). In contrast to previous work that only considered the duality between user utterances and belief states \cite{bort, dual_dst}, we discover additionally modeling the duality between user utterances and system responses further explores the implicit information in the data.
% Specifically, besides reconstructing the user utterances based on the dialogue state, the model also requires inferring the user utterances backward based on the system responses. Thus the bipartite alignment information between the data is leveraged and enhances the performance of the end-to-end dialogue system, especially in low-resource scenarios.

To address the above challenges, we propose an innovative multijugate dual learning framework in TOD (MDTOD). Contrary to previous work on reconstructing user discourse based on belief states \cite{bort, dual_dst}, we observed that modeling the duality between user utterance and system responses can further uncover alignment information of entities between user utterance, system responses, and dialogue states. Specifically, the model is required to reconstruct the user discourse based on the dialogue state and also to deduce the user utterance backward based on the system response. Consequently, the model can further learn the mapping relationship between the heterogeneous information, and improve the performance of the end-to-end TOD system in low-resource scenarios.

% However, proper dual training may enhance the likelihood of the model learning misleading correlations between the data \cite{robust_tod}, i.e., substituting the original user utterances with merely a few high-frequency phrases during training provides comparable predictions. This phenomenon reveals that the model learns superficial statistical patterns and, as a result, does not generalize well to more challenging test cases, which impacts the model's generalization.
% Meanwhile, dual learning disguisedly amplifies the correlation between user input and system responses, rendering the model susceptible to overfitting the training set and altering the response diversity \cite{cui2019dal}.

% For this purpose, we utilize a variety of semantic representations to turn the one-to-one dual learning paradigm into multijugate dual learning. Specifically, a user utterance is rephrased into multiple pairs of semantically coherent but differently formed utterance-dialog state pairings utilizing decoding methods such as beam search or sampling under the constraint of a deterministic dialogue state. On these expanded data pairings, the model then executes multijugate dual learning. 
%Consequently, the spurious correlation phenomena of the model learning superficial statistical patterns is effectively attenuated, enhancing the model's generalization performance and the diversity of its produced responses. 

However, proper dual training increases the likelihood of the model learning spurious data correlations. It is evidenced by the fact that comparable model performance can be attained using only high-frequency phrases as the training set \cite{robust_tod}. As a result, the model does not generalize well to test samples with significant expression variations or domain differences, as illustrated in Figure \ref{fig:task_intro}. To accomplish this, we expand the one-to-one dual learning paradigm to multijugate dual learning by capitalizing on the property of semantic representation variety. Given a deterministic dialog state as a constraint \cite{constrain_beam_search}, a specific user utterance (system response) is rewritten into multiple utterances (responses) with the same semantics but various expressions utilizing decoding methods such as beam search or random sampling. Consequently, the richer representation of information permits the spurious correlation of shallow statistical patterns acquired by the model to be effectively mitigated, thereby enhancing the model's generalization \cite{cui2019dal}.

% By constructing a dual training objective, our proposed technique exploits the latent information inside the data, mitigates the issue of models learning spurious correlations, and enhances the quality of responses by paraphrase-augmented multijugate relationships of the samples. As a training strategy to increase data efficiency, this method introduces no additional model parameters and may be implemented easily into any end-to-end TOD system.
% Multiple task-oriented datasets, such as MultiWOZ2.0 \cite{multiwoz}, MultiWOZ2.1 \cite{multiwoz21} and In-Car Assistant (In-Car) \cite{in-car}, are utilized to illustrate its applicability. We further demonstrate that our method of leveraging the data's underlying information provides considerable benefits in cases with limited resources \footnote{All code and parameters will be made accessible after publication.}. 

Our proposed method exploits the entity alignment information among heterogeneous data by designing a dual learning task; it also mitigates the phenomenon of false correlations and increases the generalization capacity of models via rephrase-enhanced multijugate dual learning. As a result, the method does not introduce any additional trainable model parameters. It can be directly integrated into end-to-end TOD systems in arbitrary low-resource scenarios as a training approach to increase data utilization efficiency. We show the effectiveness of our method in several task-oriented datasets, including MultiWOZ2.0 \cite{multiwoz}, MultiWOZ2.1 \cite{multiwoz21}, and KVRET \cite{kvret}. We also demonstrate the advantages of our approach in low-resource scenarios. All code and parameters will be made public.

Our primary contributions are summarized below:
\begin{itemize}
    \item A novel, model-independent, dual learning technique intended for low-resource end-to-end TOD systems is presented that can be incorporated directly into the training of any TOD system.
    \item To address the issue of spurious correlations impacting the generalization of models, a paradigm of paraphrase-enhanced multijugate dual learning is presented.
    \item We empirically evaluate the technique on several datasets, achieving competitive results without introducing extra model parameters or further pre-training and state-of-the-art results in low-resource circumstances.
\end{itemize}

\begin{figure*}[t]
\centering
\includegraphics[width=0.99\textwidth]{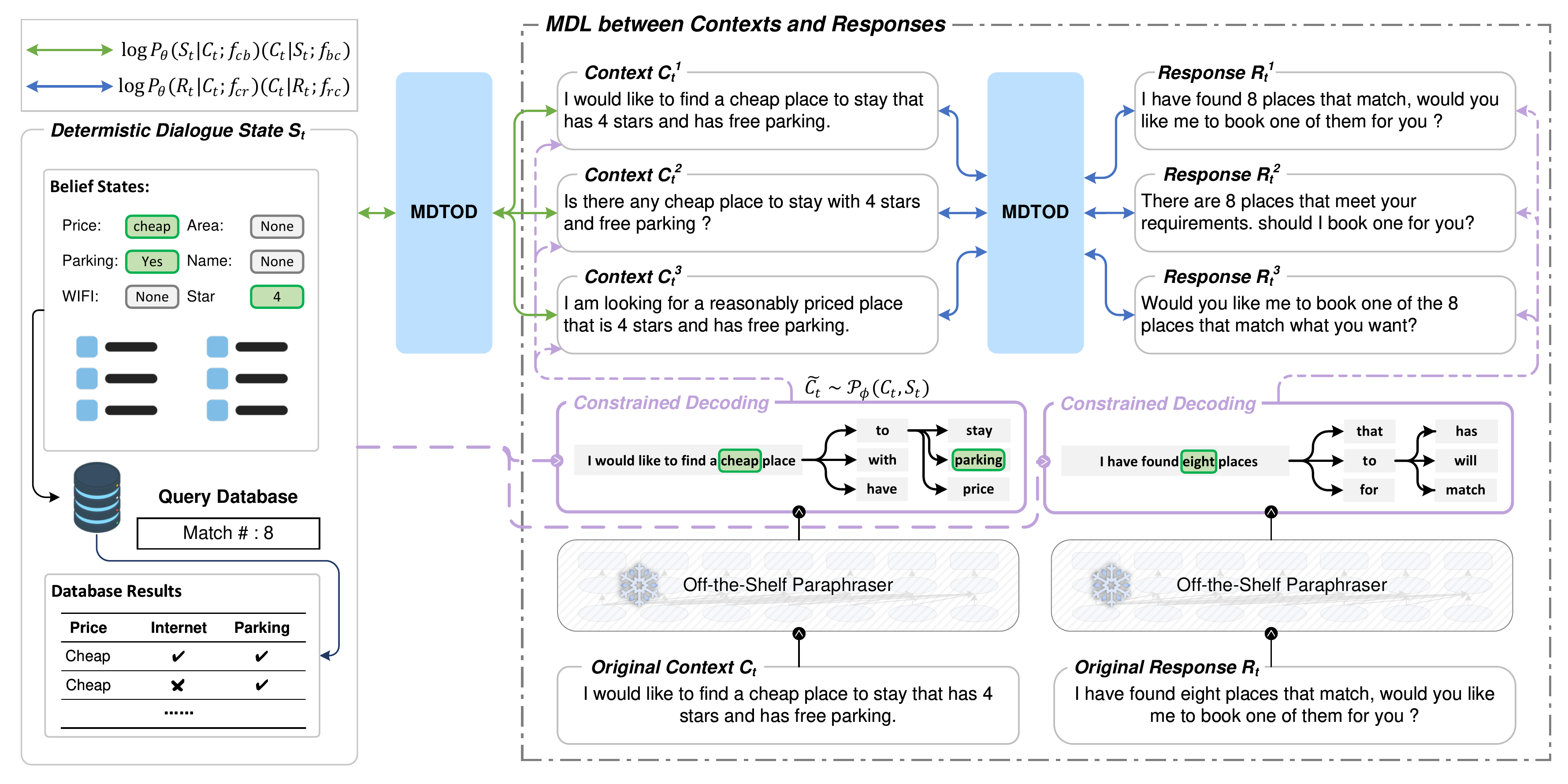} % Reduce the figure size so that it is slightly narrower than the column.
\caption{The overall structure of multijugate dual learning. To get paraphrase-enhanced multiple contexts $\tilde{C}_t$ and responses $\tilde{R}_t$, the contexts and responses in each dialogue turn will be paraphrased based on deterministic dialogue states using an off-the-shelf paraphrase model. Then, the multijugate dual learning is performed between the paraphrase-enhanced contexts $\tilde{C}_t $ and dialogue states and between the paraphrase-enhanced responses $\tilde{R}_t$ and dialogue states, respectively.}
\label{fig:main_model}
\end{figure*}

\section{Related Work}

\subsection{Task-Oriented Dialogue Systems}
TOD aims to complete user-specific goals via multiple turns of dialogue. Prior work focused mainly on TOD subtasks based on the pipeline paradigm \cite{survey_conv_ai}, but it was prone to error propagation between modules. Therefore, recent research has attempted to model dialogue tasks from an end-to-end generation approach. DAMD \cite{damd} generates the different outputs of a conversation process via multiple decoders and expands multiple dialogue actions dependent on the dialogue state. A portion of the study \cite{simpletod, ubar, soloist} models the individual dialogue tasks in the TOD as cascading generation tasks using GPT2 \cite{gpt2} of the decoder architecture as the backbone network. Multi-task approaches \cite{mintl, pptod, mttod} utilizing encoder-decoder architectures such as T5 \cite{t5} or BART \cite{bart} exist for modeling dialogue sub-tasks as sequence-to-sequence generating tasks.

% Although all the systems mentioned above employ a unified end-to-end approach to model TOD, none fully exploit the alignment information implied by heterogeneous data. 

% To this end, we anticipate that multijugate dual learning between user utterances, dialogue states, and system responses will enable better data exploitation, resulting in a more robust performance. 

Although the methods mentioned above use a uniform end-to-end approach to model TOD, none performs well in low-resource scenarios. To this end, we devise a rephrase-enhanced multijugate dual learning to exploit the entity alignment information more adequately and to obtain more robust performance.

\subsection{Dual Learning for Generation}

Dual learning aims to utilize the paired structure of data to acquire effective feedback or regularization information, thus enhancing model training performance. Dual learning was initially introduced in unsupervised machine translation \cite{dual_for_mt} and combined with reinforcement learning to optimize two agents iteratively. DSL \cite{dsl} then extended dual learning to supervised settings to take advantage of pairwise relationships of parallel corpora. Similar work \cite{guo2020cyclegt} employs cycle training to enable unsupervised mutual generation of structured graphs and text.
MPDL \cite{mpdl} expands the duality in dialogue tasks to stylized dialogue generation without the parallel corpus. A portion of the work \cite{bort,dual_dst} integrates the idea of duality into the dialogue state tracking. Some of the work \cite{dual_coherence, dual_person,cui2019dal} introduces dual learning in dialogue generation to enhance responses' diversity, personality, or coherence.
However, each method mentioned above requires multiple models or combines reinforcement learning and dual modeling, considerably increasing the task's complexity and training difficulty. 

In contrast to previous work, our proposed multijugate dual learning objectives share the same model parameters. It does not require modifications to the original training objectives of the maximum likelihood estimation, making training more straightforward and more readily applicable to other tasks.

\section{Methodology}

\subsection{End-to-End Task-Oriented Dialogue System}
Typically, end-to-end TOD systems consist of subtasks such as dialogue state prediction and response generation \cite{mttod}. End-to-end TOD systems typically model the several subtasks of the dialogue process as sequence generation tasks to facilitate the unification of model structure, and training objectives \cite{simpletod}. Denote the TOD dataset as $\mathcal{D}_\textrm{TOD}=\{{Dial_i,\textrm{DB}}\}_{i=1}^N$, where $\textrm{DB}$ is the database. In a multi-turn dialogue $Dial_i$, where the user utterance in the $t$-th turn is $U_t$, and the system response is $R_t$, the dialogue history or dialogue context can be expressed as follows:
\begin{flalign}
    C_t=[U_0, R_0, \cdots, U_{t-1}, R_{t-1}, U_t].
\end{flalign}

After that, the model generates the dialogue state $B_t$ based on the previous dialogue context $C_t$:

\begin{flalign}
    \mathcal{L}_{B}=\sum_{i=1}^N \sum_{t=1}^{T_i} -\log P_{\theta}(B_t|C_t), \label{eq:gen_dst}
\end{flalign}

\noindent where $N$ represents the total number of sessions in the dataset, $T_i$ symbolizes the total number of turns per session and $\theta$ denotes an arbitrary generation model. The system then searches the database with the criterion $B_t$ and retrieves the database result $D_t$. Then, the TOD system generate the response $R_t$ based on the context $U_t$, dialogue state $B_t$ and database query result $D_t$ for each round:

\begin{align}
    \mathcal{L}_R=\sum_{i=1}^{N}\sum_{t=1}^{T_i}-\log P_{\theta}(R_t|C_t, B_t,D_t).
\end{align}

Finally, a human-readable response text containing the entity is obtained by combining the belief state and the search results from the database.

% We combine the generated dialogue state $B_t$, the database query results $D_t$, and the system action $A_t$ with the system response $R_t$ to serve as the history in the subsequent dialogue turn:

% \begin{flalign}
%     R_t=[B_t,D_t,A_t,R_t].
% \end{flalign}

% Consequently, the following round of response generation for the current dialogue might incorporate more dialogue information. In addition, a dynamic dialogue history window technique is established to utilize the maximum number of recent dialogue history rounds without surpassing the pre-trained model's longest position encoding.

\subsection{Multijugate Dual Learning}
% Previous end-to-end dialogue systems modeled subtasks in TOD as unidirectional sequence generation tasks \cite{simpletod}, i.e., generating dialogue states and system responses based only on user utterances or dialogue contexts. The diverse alignment information between deterministic dialogue states and uncertain utterances is underutilized, according to our observations.
% Therefore, we included paraphrase-enhanced multijugate dual learning into the end-to-end TOD so that the model may acquire more relevant information implied by the data. Paraphrasing is then employed to represent the original one-to-one dual learning as a many-to-one multijugate dual learning to alleviate superficial statistical features or spurious correlations of the data that the model learned.
This section describes how to design dual learning objectives in the training process of TOD. Also, we expound on how to construct multijugate dual learning by paraphrasing user utterances and system responses with representational diversity based on deterministic dialogue states.

\subsubsection{Dual Learning in TOD}
We define the deterministic dialogue state $S_t=[B_t;D_t]$ consisting of two informational components: the belief state $B_t$ and the database query results $D_t$.

As illustrated in Figure \ref{fig:main_model}, dialogue states can be viewed as information with a unique manifestation of determinism \cite{damd} without regard to the order of dialogue actions. Utilizing dialogue state as a constraint, the natural language of context and response could be viewed as data with different representations of uncertainty. Therefore, we designed the dual task in TOD to learn the mapping relationship between the utterance of linguistic forms and dialogue state representation.
% Let $f_{cb}:C_t \rightarrow B_t,f_{bc}:B_t\rightarrow C_t$ denote the dual learning task that the dialogue context and the dialogue state mutually generate each other, $f_{cr}:(C_t,S_t)\rightarrow R_t,f_{rc}:R_t\rightarrow ( C_t,S_t)$ is the dual learning between the dialogue context $(C_t,S_t)$ and the system response $R_t$. Then, the overall objective of dual learning is:

% \begin{flalign}
%     \mathcal{L}_{\textrm{Dual}_1}=&-\mathop{\mathbb{E}}\limits_{i\sim N\atop t\sim T_i}[\log P_{\theta}(S_t^i|C_t^i;f_{cb})(C_t^i|S_t^i;f_{bc})], \\
%     \mathcal{L}_{\textrm{Dual}_2}=&-\mathop{\mathbb{E}}\limits_{i\sim N\atop t\sim T_i}[\log P_{\theta}(C_t^i,S_t^i|R_t^i;f_{rc}) \nonumber \\
%     &+\log P_{\theta}(R_t^i|C_t^i,S_t^i;f_{cr})].
% \end{flalign}
Let $f_{cb}:C_t \longmapsto B_t$ denote the forward learning objective of generating belief states according to the context referred to by Eq.\ref{eq:gen_dst}, and $f_{bc}:B_t \longmapsto C_t$ denote the reverse learning objective of reconstructing the context according to the belief states, then the dual learning task between user utterance and dialogue state is defined as maximizing the following logarithmic probability:

\begin{flalign}
    \log \sum_{i\in N}\sum_{t\in T_i}P_{\theta}(S_t^i|C_t^i;f_{cb})(C_t^i|S_t^i;f_{bc}).
\end{flalign}

Similarly, let $f_{cr}:C_t\longmapsto R_t,f_{rc}:R_t\longmapsto C_t$ denote the dual learning task between the dialogue context $C_t$ and the system response $R_t$:

\begin{flalign}
    % \mathcal{L}_{\textrm{Dual}_2}=-\mathop{\mathbb{E}}\limits_{i\sim N\atop t\sim T_i}(\log P_{\theta}(R_t^i|C_t^i;f_{cr})(C_t^i|R_t^i;f_{rc})).
    \log \sum_{i\in N}\sum_{t\in T_i}P_{\theta}(R_t^i|C_t^i;f_{cr})(C_t^i|R_t^i;f_{rc}).
\end{flalign}

Accordingly, the loss function of the total dual learning objective is the sum of the above two components:

% By learning bidirectional alignment information across heterogeneous data, the model exploits the structural knowledge inherent within the data to a greater extent. In a multi-task paradigm, it is vital to notice that the four mapping tasks share a set of model parameters to keep the setting of end-to-end TOD processes accomplished utilizing a single model.

\begin{flalign}
    &\mathcal{L}_\textrm{Dual}=\mathop{\mathbb{E}}\limits_{\substack{i\sim N\\ t\sim T_i}}-(\log P_{\theta}(S_t^i,R_t^i|C_t^i;f_{cr},f_{cb})\nonumber\\
    &+\log P_\theta(C_t^i|S_t^i;f_{bc})+\log P_\theta(C_t^i|R_t^i;f_{rc})).
\end{flalign}

Furthermore, the two dual learning objectives share a set of model parameters in a multi-task paradigm, thus ensuring knowledge transfer between the dual tasks.

\subsubsection{Construction of Multijugate Relations}
Dual learning enhances data usage efficiency by acquiring additional entity alignment information between heterogeneous data, but it does not lessen the effect of spurious correlations on model generalization. Leveraging the deterministic properties of dialogue states and the uncertainty of linguistic representations, we expand the original one-to-one dual learning to multijugate dual learning by paraphrases.
% In particular, a given dialogue state $S_t$ in a $t$-round initially correlates to a context $C_t$ or a system response $R_t$, producing the component data pair $(S_t, C_t)$ or $(S_t, R_t)$, where $C_t, R_t$ are constrained to $S_t$.
Theoretically, several semantically identical but inconsistently expressed contexts or system responses exist for a deterministic dialogue state. Consequently, given $(S_t, C_t)$ or $(S_t, R_t)$, we rephrase the context $C_t$ and the response $R_t$ restricted by the entities in dialogue state $S_t$ with the following constraint generation method:

\begin{align}
    \tilde{C}_t\sim\mathcal{P}(C_t, S_t), \tilde{R}_t\sim \mathcal{P}(S_t, R_t).
\end{align}

Specifically, we utilize an off-the-shelf paraphrasing model with the dialogue context $C_t$ as the model input. Also the value in the dialogue state $S_t$ will be treated as a constraint to limit the decoding. Then, beam search is employed in the generation to obtain $K$ different contexts $\tilde{C}_t$ or responses $\tilde{R}_t$ as the result of paraphrase generation.

Moreover, since the context $C_t$ of the current turn depends on the dialogue history $(\cdots, C_{t-1}, S_{t-1}, R_{t-1})$ of the previous turn, rewriting the context or responses of each turn results in a combinatorial explosion. Therefore, a heuristic was adopted whereby the dialogue context $C_t$ and system response $R_t$ would only be rewritten once every dialogue turns. The method for producing the final paraphrase is:

\begin{align}
    \tilde{C}_t^{ij}&\sim\sum_{i=1}^N\sum_{t=1}^{T_i}\sum_{j=1}^M \mathcal{P}(C_t^{ij},S_t^{ij}),\\
    \tilde{R}_t^{ij}&\sim\sum_{i=1}^N\sum_{t=1}^{T_i}\sum_{j=1}^M \mathcal{P}(S_t^{ij},R_t^{ij}),
\end{align}

\noindent where $M$ represents the number of single samples to be rewritten. In practice, as the proportion of training data increases, the number of $M$ decreases. In addition, paraphrasing was preferred over word substitution or addition/deletion-based techniques \cite{eda} because word substitution is based on a particular probability of word-level alterations, preventing the modification of phrases with false correlation. Moreover, section \ref{sec:abla_para} approved paraphrasing produces more diverse and high-quality augmented content, alleviating the risk of spurious relevance more effectively.

\subsubsection{Multijugate Dual Learning for Training}
By acquiring paraphrase-enhanced samples, the original one-to-one dual learning can be augmented with multijugate dual learning, allowing the model to completely leverage the entity alignment information between heterogeneous data while maintaining appropriate generalization. The overall framework of our method is illustrated in Figure \ref{fig:main_model}. Consequently, the final loss function for multijugate dual learning of TOD is as follows:

\begin{align}
    \tilde{\mathcal{L}}_\textrm{Dual}&=\mathop{\mathbb{E}}\limits_{\substack{i\sim N\\ t\sim T_i\\ j\sim M}}-(\log P_{\theta}(S_t^{ij},R_t^{ij}|C_t^{ij};f_{cr},f_{cb})\nonumber\\
    &+\log P_\theta(C_t^{ij}|S_t^{ij};f_{bc})(C_t^{ij}|R_t^{ij};f_{rc})).
\end{align}

% The process of generating a dialogue state or response from context is termed  \textit{forward}, and vice versa. The total of the two dual learning objectives constitutes the overall objective during the training phase:

% \begin{align}
%     \mathcal{L}=\tilde{\mathcal{L}}_{\textrm{Dual}_1}+\lambda \tilde{\mathcal{L}}_{\textrm{Dual}_2},
% \end{align}

% \noindent where $\lambda$ is the coefficient controlling the magnitude of the loss.

\begin{table*}[t]
\centering
\resizebox{1.0\textwidth}{!}{
\setlength{\tabcolsep}{1.2mm}
\begin{tabular}{lllllllllllll}
\toprule
             & \multicolumn{12}{c}{\textbf{MultiWOZ 2.0}}                                                                                                                 \\
\cmidrule(lr){2-5}\cmidrule(lr){6-9}\cmidrule(lr){10-13}
             & \multicolumn{4}{c}{5\% Training set}                & \multicolumn{4}{c}{10\% Training set}        & \multicolumn{4}{c}{20\% Training set}        \\
\midrule
\textbf{Model}        & Inform & Success        & BLEU           & Comb. & Inform & Success & BLEU           & Comb. & Inform & Success & BLEU           & Comb. \\
\midrule
MD-Sequicity & 49.40  & 19.70          & 10.30          & 44.85    & 58.10  & 34.70   & 11.40          & 57.80    & 64.40  & 42.10   & 13.00          & 66.25    \\
DAMD         & 52.50  & 31.80          & 11.60          & 53.75    & 55.30  & 30.30   & 13.00          & 55.80    & 62.60  & 44.10   & 14.90          & 68.25    \\
SOLOIST      & 69.30  & 52.30          & 11.80          & 72.60    & 69.90  & 51.90   & 14.60          & 75.50    & 74.00  & 60.10   & 15.25          & 82.29    \\
MinTL        & 75.48  & 60.96          & 13.98          & 82.20    & 78.08  & 66.87   & 15.46          & 87.94    & 82.48  & 68.57   & 13.00          & 88.53    \\
UBAR         & 73.04  & 60.28          & \textbf{16.03} & 82.89    & 79.20  & 68.70   & 16.09          & 90.04    & 82.50  & 66.60   & \textbf{17.72} & 92.26    \\
T5-Base      & 77.80  & 63.30          & 14.56          & 84.94    & 81.00  & 67.00   & 15.17          & 89.17    & 84.20  & 72.70   & 17.71          & 96.16    \\
BORT         & 69.80  & 45.90          & 11.00          & 68.90    & 74.50  & 60.60   & 15.50          & 83.10    & 82.10  & 65.60   & 14.30          & 88.10    \\
PPTOD        & 79.86  & 63.48          & 14.89          & 86.55    & 84.42  & 68.36   & 15.57          & 91.96    & 84.94  & 71.70   & 17.01          & 95.32    \\
MTTOD        & 82.00  & \textbf{64.00} & 14.48          & 87.49    & 82.10  & 71.10   & \textbf{16.21} & 92.81    & 89.50  & 78.50   & 15.53          & 99.53    \\
\midrule
MDTOD &
  \begin{tabular}[c]{@{}l@{}}\textbf{85.65} \\ ($\pm$2.35)\end{tabular} &
  \begin{tabular}[c]{@{}l@{}}62.20 \\ ($\pm$2.70)\end{tabular} &
  \begin{tabular}[c]{@{}l@{}}15.24 \\ ($\pm$1.04)\end{tabular} &
  \begin{tabular}[c]{@{}l@{}}\textbf{89.16} \\ ($\pm$1.48)\end{tabular} &
  \begin{tabular}[c]{@{}l@{}}\textbf{86.30} \\ ($\pm$0.90)\end{tabular} &
  \begin{tabular}[c]{@{}l@{}}\textbf{71.50} \\ ($\pm$0.60)\end{tabular} &
  \begin{tabular}[c]{@{}l@{}}14.47 \\ ($\pm$1.19)\end{tabular} &
  \begin{tabular}[c]{@{}l@{}}\textbf{93.37} \\ ($\pm$1.04)\end{tabular} &
  \begin{tabular}[c]{@{}l@{}}\textbf{90.25} \\ ($\pm$0.55)\end{tabular} &
  \begin{tabular}[c]{@{}l@{}}\textbf{80.90} \\ ($\pm$0.42)\end{tabular} &
  \begin{tabular}[c]{@{}l@{}}16.40 \\ ($\pm$1.15)\end{tabular} &
  \begin{tabular}[c]{@{}l@{}}\textbf{101.97} \\ ($\pm$0.73)\end{tabular} \\
\bottomrule
\end{tabular}
}
\caption{The performance of MDTOD is evaluated at 5\%, 10\%, and 20\% of the data size. Comb. denotes Combined Score.}
\label{tab:fewshot_results}
\end{table*}

\section{Experiments}

In the context of an end-to-end dialogue scenario, we examine the comprehensive performance of multijugate dual learning on several dialogue datasets, including performance on dialogue state tracking and end-to-end task completion. In addition, evaluation studies were conducted in a scenario with limited resources to assess how effectively dual learning utilizes the knowledge contained within the data. In addition, the impact of several dual learning components and rewriting procedures on the method's overall performance is investigated.

\subsection{Datasets and Evaluation Metrics}
MultiWOZ2.0 \cite{multiwoz}, MultiWOZ2.1 \cite{multiwoz21}, and KVRET \cite{kvret}, three of the most extensively investigated datasets in the task-oriented dialogue domain, were analyzed. MultiWOZ2.0 is the first proposed dialogues dataset across seven domains, and MultiWOZ2.1 is the version with several MultiWOZ2.0 annotation problems fixed. Following earlier research, we simultaneously evaluate both datasets to assess the robustness of the model against mislabeling.
KVRET is a multi-turn TOD dataset containing three domains: calendar scheduling, weather query, and navigation. Detailed statistics of the three datasets are illustrated in Table \ref{tab:data_statics}.

For the selection of metrics under the end-to-end dialogue task, we use the standard and widely used \texttt{Inform}, \texttt{Success}, \texttt{BLEU}, and \texttt{Combined score}, where \texttt{Inform} measures whether the system's responses refer to the entity requested by the user, \texttt{Success} measures whether the system has answered all of the user's requests, \texttt{BLEU} measures the quality of the model generation. The \texttt{Combined score} indicates the overall performance of the task-oriented system. It is calculated using the formula: \texttt{Combined Score = (Inform + Success) * 0.5 + BLEU}. For the dialogue state tracking task, the Joint Goal Accuracy (JGA) is applied to quantify the fraction of total turns where the model predicts that all slots in one turn are correct.

\subsection{Baselines}
We did comparison experiments with the following potent baselines. (1) \textbf{DAMD} \cite{damd}: addresses the one-to-many issue in dialogue by extending dialogue states to many system actions. (2) \textbf{SimpleTOD} \cite{simpletod}: A language model serves as the foundation for end-to-end TOD tasks by generating sequential dialogue states, dialogue actions, and dialogue responses. (3) \textbf{DoTS} \cite{dots}: tackles the problem of higher memory consumption owing to lengthy conversation histories by reducing the context and adding domain states as contexts. (4) \textbf{SOLOIST} \cite{soloist}: further pre-training on heterogeneous dialogue data and transfer learning for dialogue tasks downstream. (5) \textbf{MinTL}: employs a copy method to carry over past dialogue states and introduces Levenshtein belief spans to generate a minimal amount of dialogue states efficiently. (6) \textbf{UBAR} \cite{ubar}: considers belief states, system actions, and system responses as dialogue contexts, hence optimizing the utilization of the dataset's content. (7) \textbf{PPTOD} \cite{pptod}: A T5-based backbone network with additional pre-training on numerous dialogue datasets and simultaneous multitasking of several dialogue tasks with prompt learning. (8) \textbf{MTTOD} \cite{mttod}: Using T5 as the backbone model, two decoders were employed to create dialogue states and system responses, and an additional span prediction task was introduced on the encoder side. (9) \textbf{BORT} \cite{bort}: utilizing denoised reconstruction to recover noisy dialogue states and system responses.

\subsection{Overall Results}

\begin{table*}[t]
\centering
\resizebox{1.0\textwidth}{!}{
\setlength{\tabcolsep}{1.2mm}
\begin{tabular}{lcccccccccccc}
\toprule
 & \multicolumn{12}{c}{\textbf{KVRET}} \\
 \cmidrule(lr){2-5}\cmidrule(lr){6-9}\cmidrule(lr){10-13}
 & \multicolumn{4}{c}{10\% Training set} & \multicolumn{4}{c}{20\% Training set} & \multicolumn{4}{c}{50\% Training set} \\
\midrule 
 & Inform   & Success   & BLEU  & Comb.  & Inform   & Success   & BLEU  & Comb.  & Inform   & Success   & BLEU  & Comb.  \\
\midrule
T5 &
  \begin{tabular}[c]{@{}c@{}}75.82\\ (3.42)\end{tabular} &
  \begin{tabular}[c]{@{}c@{}}18.30\\ (6.74)\end{tabular} &
  \begin{tabular}[c]{@{}c@{}}10.51\\ (0.77)\end{tabular} &
  \begin{tabular}[c]{@{}c@{}}57.57\\ (5.14)\end{tabular} &
  \begin{tabular}[c]{@{}c@{}}80.25\\ (3.08)\end{tabular} &
  \begin{tabular}[c]{@{}c@{}}50.81\\ (8.71)\end{tabular} &
  \begin{tabular}[c]{@{}c@{}}15.72\\ (1.75)\end{tabular} &
  \begin{tabular}[c]{@{}c@{}}81.25\\ (6.26)\end{tabular} &
  \begin{tabular}[c]{@{}c@{}}83.42\\ (2.57)\end{tabular} &
  \begin{tabular}[c]{@{}c@{}}70.45\\ (3.13)\end{tabular} &
  \begin{tabular}[c]{@{}c@{}}17.26\\ (1.27)\end{tabular} &
  \begin{tabular}[c]{@{}c@{}}94.20\\ (2.15)\end{tabular} \\
T5+DL &
  \begin{tabular}[c]{@{}c@{}}73.82\\ (1.29)\end{tabular} &
  \begin{tabular}[c]{@{}c@{}}33.11\\ (9.10)\end{tabular} &
  \begin{tabular}[c]{@{}c@{}}11.55\\ (1.53)\end{tabular} &
  \begin{tabular}[c]{@{}c@{}}65.02\\ (6.36)\end{tabular} &
  \begin{tabular}[c]{@{}c@{}}\textbf{82.25}\\ (0.68)\end{tabular} &
  \begin{tabular}[c]{@{}c@{}}59.58\\ (3.76)\end{tabular} &
  \begin{tabular}[c]{@{}c@{}}16.18\\ (0.90)\end{tabular} &
  \begin{tabular}[c]{@{}c@{}}87.09\\ (2.62)\end{tabular} &
  \begin{tabular}[c]{@{}c@{}}81.07\\ (5.16)\end{tabular} &
  \begin{tabular}[c]{@{}c@{}}\textbf{74.05}\\ (1.18)\end{tabular} &
  \begin{tabular}[c]{@{}c@{}}18.59\\ (0.90)\end{tabular} &
  \begin{tabular}[c]{@{}c@{}}96.15\\ (2.94)\end{tabular} \\
\midrule
MDTOD &
  \begin{tabular}[c]{@{}c@{}}\textbf{78.89}\\ ($\pm$0.94)\end{tabular} &
  \begin{tabular}[c]{@{}c@{}}\textbf{56.49}\\ ($\pm$4.62)\end{tabular} &
  \begin{tabular}[c]{@{}c@{}}\textbf{14.60}\\ ($\pm$0.99)\end{tabular} &
  \begin{tabular}[c]{@{}c@{}}\textbf{82.30}\\ ($\pm$2.97)\end{tabular} &
  \begin{tabular}[c]{@{}c@{}}78.71\\ ($\pm$3.36)\end{tabular} &
  \begin{tabular}[c]{@{}c@{}}\textbf{64.03}\\ ($\pm$6.36)\end{tabular} &
  \begin{tabular}[c]{@{}c@{}}\textbf{16.57}\\ ($\pm$0.64)\end{tabular} &
  \begin{tabular}[c]{@{}c@{}}\textbf{87.94}\\ ($\pm$4.98)\end{tabular} &
  \begin{tabular}[c]{@{}c@{}}\textbf{84.15}\\ ($\pm$1.97)\end{tabular} &
  \begin{tabular}[c]{@{}c@{}}71.80\\ ($\pm$2.44)\end{tabular} &
  \begin{tabular}[c]{@{}c@{}}\textbf{19.06}\\ ($\pm$0.79)\end{tabular} &
  \begin{tabular}[c]{@{}c@{}}\textbf{97.03}\\ ($\pm$1.45)\end{tabular} \\
\bottomrule
\end{tabular}
}
\caption{The performance is evaluated at 10\%, 20\%, and 50\% of the data size. The numbers in parentheses indicate the variance of the four runs.}
\label{tab:fewshot_incar}
\end{table*}

\subsubsection{Performance in Low-resource Setting}

\textbf{MultiWOZ}  To investigate the generalizability of multijugate dual learning with limited resources, we assessed the model on the MultiWOZ2.0 dataset for dialogue sizes of 5\%, 10\%, and 20\%. As shown in Table \ref{tab:fewshot_results}, MDTOD received the highest combined score compared to baselines for all data sizes. MDTOD obtains a 1.67-point improvement in the combined score at 5\% of the training data compared to the previous best result. Our strategy produces the highest results for Inform and Success, which are task completion metrics, when applied to 10\% and 20\% of the data, respectively. In addition, our method obtains highly competitive results compared to PPTOD with additional dialogue data for pre-training and MTTOD with 50\% more parameters. Thus, the results above imply that paraphrasing augmented multijugate dual learning that leverages implicit information embedded within the data is more effective in settings with limited resources.

\noindent\textbf{KVRET}  We also evaluate the impact of multijugate dual learning on the performance improvement of TOD on the KVRET dataset. We use T5-base as the backbone network, where T5+DL indicates the addition of dual learning on T5 and MDTOD indicates the combination of multijugate dual learning on T5. From the experimental results in Table \ref{tab:fewshot_incar}, it can be concluded that after applying the dual learning objective under the low resource setting, the model achieves a significant improvement in \texttt{Success} when given different proportions of training samples, indicating that the dual learning can further learn the alignment information between entities and thus improve the success rate of the task. Meanwhile, T5+DL achieves higher values on BLEU with different proportions of training data, indicating that the dual learning objective between user utterance and system response is also beneficial for improving the quality of text generation. In addition, MDTOD with multijugate dual learning achieves better results, indicating that controlled rephrasing can further enhance the effect of dual learning.

\subsubsection{Dual Learning in Dialogue State Tracking}

\begin{table}[t]
\centering
\resizebox{1\columnwidth}{!}{
\begin{tabular}{lcccc}
\toprule
\multicolumn{1}{c}{\multirow{2}{*}{Model}} & \multicolumn{4}{c}{Training Set}                      \\
\cmidrule(lr){2-5}
\multicolumn{1}{c}{}                       & 1\%         & 5\%         & 10\%        & 20\%        \\
\midrule
SimpleTOD                                  & 7.91$_{\pm1.07}$  & 16.14$_{\pm1.48}$ & 22.37$_{\pm1.17}$ & 31.22$_{\pm2.32}$ \\
MinTL                                      & 9.25$_{\pm2.33}$  & 21.28$_{\pm1.94}$ & 30.32$_{\pm2.14}$ & 35.96$_{\pm1.25}$ \\
SOLOIST                                    & 13.21$_{\pm1.97}$ & 26.53$_{\pm1.62}$ & 32.42$_{\pm1.13}$ & 38.68$_{\pm0.98}$ \\
PPTOD$_{base}$                                & \textbf{29.72}$_{\pm0.61}$ & 40.20$_{\pm0.39}$ & 43.35$_{\pm0.64}$ & 46.96$_{\pm0.40}$ \\
\midrule
MDTOD                                      & 21.22$_{\pm2.86}$ & \textbf{40.90}$_{\pm0.20}$ & \textbf{45.10}$_{\pm1.40}$ & \textbf{47.89}$_{\pm0.55}$  \\
\bottomrule
\end{tabular}
}
\caption{DST evaluated at different proportions of low resources. The results are the means and standard deviations of the four runs.}
\label{tab:fewshot_dst}
\end{table}

To further investigate the effectiveness of the dual learning task between user utterance and dialogue state on the gain of TOD in multijugate dual learning, we conducted experiments on the MultiWOZ2.0 dataset for dialogue state tracking in low-resource scenarios. We set four different quantitative training sizes of 1\%, 5\%, 10\% and 20\% to represent different degrees of low-resource scenarios.

We can infer from the experimental results in Table \ref{tab:fewshot_dst} that MDTOD had the greatest accuracy at three different magnitudes, 5\%, 10\%, and 20\%. MDTOD is lower than PPTOD at 1\% magnitude due to that PPTOD performs further pre-training on a large amount of additional dialogue data and thus can achieve relatively better results in extremely low-resource scenarios. Conversely, MDTOD does not perform any additional pre-training, but still achieves the highest accuracy in the case of the other three magnitudes of data, indicating that multijugate dual learning between user utterances and dialogue states is an important component that makes the overall approach effective.

\subsection{Analysis}

\subsubsection{Dismantling multijugate dual learning}

\begin{table}[t]
\centering
\resizebox{1\columnwidth}{!}{
\begin{tabular}{lcccc}
\toprule
             & \multicolumn{4}{c}{MultiWOZ 2.0} \\
\cmidrule(lr){2-5}
Model        & Inform     & Success    & BLEU     & Comb.           \\
\midrule
Full         & 85.27      & 71.07      & 15.26    & 93.43           \\
-w/o Para    & 85.12      & 70.93      & 15.09    & 93.12 ($\downarrow$0.31)    \\
-w/o DU-DL   & 85.23      & 71.23      & 13.48    & 91.71 ($\downarrow$1.72)    \\
-w/o RU-DL   & 84.70      & 70.70      & 13.86    & 91.56 ($\downarrow$1.87)    \\
-w/o Both-DL & 83.20      & 70.80      & 14.42    & 91.41 ($\downarrow$\textbf{2.02})    \\
\bottomrule
\end{tabular}
}
\caption{Different setting of multijugate dual learning.}
\label{tab:ablation}
\end{table}

To investigate the effect of different dual learning components and paraphrase augmentation on the proposed technique, we conducted ablation experiments by omitting various components using a 10\% data size setting. In Table \ref{tab:ablation}, Para represents the approach of paraphrase augmentation, DU-DL represents dual learning between context and dialogue state, and RU-DL indicates dual learning between context and system response.

As shown in Table \ref{tab:ablation}, the model's performance decreases slightly when only dual learning is retained and the paraphrase enhancement is removed, indicating that multijugate dual learning can partially mitigate the overfitting problem caused by pairwise learning and thereby improve the model's generalization capability. Among the various dual learning components, removing dual learning between context and system responses resulted in a 1.87-point performance decrease, indicating that fully exploiting the implicit alignment information between context and system responses was more effective at enhancing the model's overall performance. Additionally, deleting both dual learning components resulted in a 2.02 points decrease in the combined score, demonstrating that both dual learning objectives are effective for this strategy.

\subsubsection{Mitigating Spurious Correlation for Generalization}

\begin{table}[t]
\centering
\resizebox{1\columnwidth}{!}{
\begin{tabular}{lcccc}
\toprule
          & \multicolumn{4}{c}{\textbf{KVRET}}        \\
\cmidrule(lr){2-3} \cmidrule(lr){4-5}
\textbf{Domains} & \multicolumn{2}{c}{X$_{\textrm{/schedule}}\rightarrow$schedule} & \multicolumn{2}{c}{X$_{\textrm{/weather}}\rightarrow$weather} \\
\midrule
Para. Num & Goal Score  & BLEU        & Goal Score  & BLEU       \\
\midrule
0         & 25.84$_{1.63}$ & 10.59$_{0.05}$ & 10.88$_{2.01}$ & 5.80$_{0.68}$ \\
1         & 26.26$_{1.17}$ & 10.01$_{0.50}$ & 13.40$_{3.59}$ & 5.02$_{0.05}$ \\
2         &\textbf{ 26.70}$_{0.72}$ & \textbf{11.30}$_{1.05}$ & \textbf{15.09}$_{2.29}$ & \textbf{5.88}$_{0.37}$ \\
\bottomrule
\end{tabular}
}
\caption{The outcomes of the cross-domain evaluation. X$_{/\ast}\rightarrow \ast$ denotes that the $\ast$ domain is excluded from the training set and only the $\ast$ domain is tested.}
\label{tab:para_for_gen}
\end{table}

This section explores the generalizability of dual learning across domains when different numbers of paraphrases are tested, i.e., on a domain that does not appear in the training process, to examine the effect of rephrasing enhanced multijugate dual learning for mitigating spurious correlations of entities and improving generalization. In the In-Car dataset, we explore the ability of MDTOD to generalize to both the scheduling and weather domains separately.

The \texttt{Goal Score} is calculated as \texttt{(inform + success) * 0.5} to signify task accomplishment. As indicated in Table \ref{tab:para_for_gen}, the model exhibits some improvement in task completion rate and text generation performance in both new domains when using rephrased augmented multijugate dual  learning. Further, when the number of paraphrases is 2, a boost of 4.21 points is obtained on the \texttt{Goal Score} compared to no additional rephrasing mechanism. This improvement indicates that the multiple conjugations further alleviate the shallow spurious correlations among entities captured by the model, thus improving the task completion rate.

\subsubsection{Effect of Different Paraphrases}
\label{sec:abla_para} 

\begin{figure}[t]
\centering
\includegraphics[width=0.98\columnwidth]{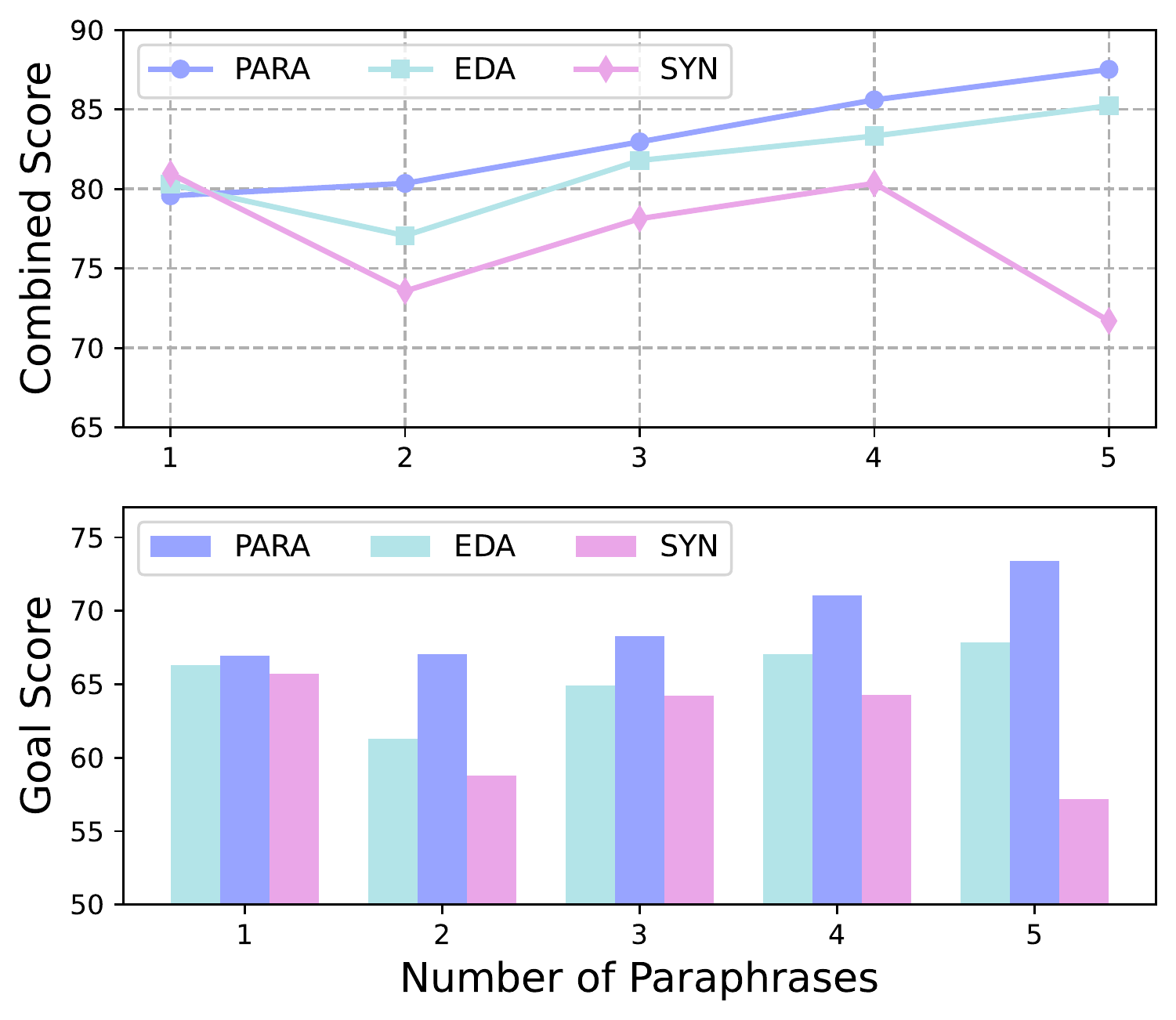} % Reduce the figure size so that it is slightly narrower than the column. Don't use precise values for figure width.This setup will avoid overfull boxes.
\caption{To investigate the impact of various rephrasing strategies on multijugate dual learning.}
\label{fig:para_analysis}
\end{figure}

To investigate the impact of various rephrasing techniques on the construction of multijugate dual learning, we examined the impact of easy data augmentation (EDA) \cite{eda}, synonym replacement (SYN), and paraphrasing (PARA) to generate augmented data with limited resources. As demonstrated in the upper part of Figure \ref{fig:para_analysis}, both PARA and EDA demonstrate minor improvements as the number of augmented data increases, with PARA exceeding EDA. The results indicate that PARA generates higher-quality augmented data, whereas SYN increases noise.

The results in Figure \ref{fig:para_analysis} indicate that increasing the number of PARA leads to an increase in the completion rate of dialogue goals. In contrast, EDA and SYN provide a minor boost or decrease in the model's performance. This analysis reveals that a rephrasing strategy enables better discourse rewriting under dialogue state constraints, alleviating the spurious correlation issue and enhancing the model's generalizability.

\section{Conclusion}
% We propose a novel multijugate dual learning architecture for task-oriented dialogues, facilitating the full exploitation of heterogeneous implicit alignment information in the data and mitigating the spurious correlation of shallow pattern statistics. Experiments on several datasets demonstrate that the proposed method delivers state-of-the-art performance in dialogue state tracking and end-to-end response generation, particularly in low-resource circumstances.
% Meanwhile, our work has scope for future investigation, including how to employ more thoroughly developed rephrasing tactics, qualitative filtering of augmented texts, and construct adequate dual learning objectives.

We propose a novel multijugate dual learning for task-oriented dialogues in low-resource scenarios. Exploiting the duality between deterministic dialogue states and uncertain utterances enables the entity alignment information in heterogeneous data to be fully exploited. Meanwhile, paraphrase-enhanced multijugate dual learning alleviates the spurious correlation of shallow pattern statistics. Experiments on several TOD datasets show that the proposed method achieves state-of-the-art results in both end-to-end response generation and dialogue state tracking in low-resource scenarios.

\section*{Limitations}

Multijugate dual learning improves the model's performance in TOD tasks in low-resource scenarios, but the introduction of the dual training objects increases the required graphics memory and training steps. In addition, the rephrasing mechanism necessitates an additional paraphraser to rewrite the training samples; hence, the number of training samples increases according to the number of paraphrases. Despite this, we find that the higher training cost associated with multijugate dual learning is preferable to employing a large quantity of dialogue data for further pre-training or manually labeling data.

Considered from a different angle, the scenario described above presents possibilities for future research, such as the development of higher-quality rephrasing algorithms to filter the augmented text. In the meantime, multijugate dual learning is a learning objective between structured and unstructured texts. Therefore it may be extended to any task involving heterogeneous data, such as generative information extraction, and data-to-set generation.

\section*{Acknowledgements}

This work was supported by the National Key Research and Development Program of China (No.2020AAA0108700) and National Natural Science Foundation of China (No.62022027).

\bibliography{main}
\bibliographystyle{acl_natbib}

\clearpage
\appendix

\section{Implementation Details}
\label{sec:appendix}

\begin{table}[t]
\centering
\resizebox{0.95\columnwidth}{!}{
\begin{tabular}{lccc}
\toprule
\textbf{Parameters} & \textbf{MultiWOZ2.0} & \textbf{MultiWOZ2.1} & \textbf{KVRET} \\
\midrule
Optimizer           & AdamW                & AdamW                & AdamW           \\
LR Scheduler        & Linear               & Linear               & Linear          \\
LR                  & \multicolumn{2}{c}{\{2e-4,4e-4, 6e-4, 8e-4\}}  & \{6e-4, 8e-4\}               \\
Warmup ratio        & 0.2                  & 0.2                  & \{0.2, 0.3, 0.4\}             \\
Epoch               & 10                   & 10                   & 6               \\
Top-p               & 0.7                  & 0.7                  & 0.7             \\
Input Length        & 512                  & 512                  & 512             \\
Output Length       & 200                  & 200                  & 200             \\
\bottomrule
\end{tabular}
}
\caption{Hyper-parameters used for MultiWOZ2.0, MultiWOZ2.1 and In-Car.}
\label{tab:hyper-para}
\end{table}

\subsection{Setup for Experiments}
All of our experiments utilize Huggingface's checkpoints. The backbone network of the end-to-end dialogue model is \texttt{T5-base}. For the generation of paraphrases, we adopt \texttt{tuner007/pegasus\_paraphrase}\footnote{\url{https://huggingface.co/tuner007/pegasus_paraphrase}.} directly and construct multiple paraphrases with beam search in decoding.
The AdamW optimizer was applied to train the dialogue model and adjusted using linear scheduling with a warmup technique. For the entire dataset in MultiWOZ, we trained 10 epochs with a batch size of 3. Training epochs were relatively increased in the scenario with limited resources. All trials were executed on NVIDIA GeForce RTX 3090 GPU (24G) or NVIDIA A800 (80G). Without additional specifications, the average of three runs with different random seeds was taken as the final result for all experiments.

\begin{table}[t]
\centering
\resizebox{0.95\columnwidth}{!}{
\begin{tabular}{lccc}
\toprule
\textbf{Metric}    & \textbf{MWOZ2.0} & \textbf{MWOZ2.1} & \textbf{KVRET} \\
\midrule
Train & 8438                 & 8438                 & 2425            \\
Dev   & 1000                 & 1000                 & 302             \\
Test  & 1000                 & 1000                 & 304             \\
Avg. \#turns per dialog  & 13.46                 & 13.46                 & 5.25             \\
Avg. \#tokens per turn   & 13.13                 & 13.13                 & 8.02             \\
\bottomrule
\end{tabular}
}
\caption{Statistics of evaluated datasets.}
\label{tab:data_statics}
\end{table}

\section{Experiments with Full Training Data}
\subsection{End-to-End Evaluation}
Table \ref{tab:main_results} demonstrates that, given the entire dataset, our proposed technique beats the comparable baselines on both datasets. The combined score on MlutiWOZ2.1 has increased by 1.48 points compared to the previous highest result. Notably, our approach does not use more dialogue data for further pre-training, nor does it introduce additional parameters or use a more powerful pre-training model for dialogue. 
Despite this, Dual-Dialog earns the highest results, proving that dual learning can more thoroughly exploit the information included in the original data and enhance the performance of task-oriented dialogue systems despite the vast amount of data. Our proposed strategy likewise achieves the greatest BLEU on MultiWOZ2.0, showing that the quality of the model's generated responses has been substantially enhanced.

\subsection{Dialogue State Tracking}
To further investigate the influence of bipartite modeling between uncertain user utterances and deterministic belief states in dual learning on TOD systems, we compared MDTOD with different generating paradigm baselines while performing the belief state tracking task. According to Table \ref{tab:dst}, MDTOD obtained up-to-date results for both datasets in the belief state tracking challenge. On MultiWOZ 2.0 and 2.1, our suggested technique achieves a 0.41 JGA improvement above the previous highest BORT and MTTOD. Dual learning between dialogue states and user utterances can learn entity alignment information in the data, resulting in improved performance in belief state tracking.

\begin{table}[t]
\centering
\resizebox{1\columnwidth}{!}{
\begin{tabular}{lcc}
\toprule
\multicolumn{3}{c}{\textit{Generation-based Methods}}     \\
\midrule
             & \multicolumn{2}{c}{Joint Goal Accuracy} \\
                
\cmidrule(lr){2-3}
Model        & 2.0       & 2.1       \\
\midrule
TRADE \cite{trade}        & 48.62                     & 46.00                     \\
COMER \cite{comer}       & 48.79                     & -                         \\
DSTQA \cite{dstqa}       & 51.44                     & 51.17                     \\
SOM-DST \cite{somdst}     & 51.38                     & 52.57                     \\
dual-DST \cite{dual_dst}   & -                       & 49.88                    \\
T5-Base \cite{t5}      & 52.16                     & 52.08                     \\
SimpleTOD$\dagger$ \cite{simpletod}     & 51.37                 & 50.14                      \\
SOLOIST$\dagger$ \cite{soloist}     & 53.20                     & 53.36                     \\
PPTOD$\dagger$ \cite{pptod}        & 53.57                     &  51.68                    \\
MTTOD \cite{mttod}        & 53.56                     & 53.44                     \\
BORT \cite{bort}        & 54.00                     & -                         \\
\midrule
MDTOD & \textbf{54.41}            & \textbf{53.85}            \\   
\bottomrule
\end{tabular}
}
\caption{Results of the performance comparison between MDTOD and other generative models, using MultiWOZ 2.0 and 2.1 datasets, for the dialogue state tracking. $\dagger$: The results provided in the publications of these approaches could not be reproduced in MultiWOZ2.1 or with an unfair evaluation script, so we corrected these results based on their open source code.}
\label{tab:dst}
\end{table}

\begin{table*}[t]
\centering
\resizebox{0.98\textwidth}{!}{
\setlength{\tabcolsep}{1.2mm}
\begin{tabular}{lcccccccc}
\toprule
             & \multicolumn{4}{c}{\textbf{MultiWOZ 2.0}}                                   & \multicolumn{4}{c}{\textbf{MultiWOZ 2.1}}                                   \\
\cmidrule(lr){2-5}\cmidrule(lr){6-9}
\textbf{Model}        & Inform         & Success        & BLEU           & Comb.        & Inform         & Success        & BLEU           & Comb.        \\
\midrule
DAMD \cite{damd}         & 76.33          & 60.40          & 16.60          & 84.97           & -              & -              & -              & -               \\
SimpleTOD \cite{simpletod}   & 84.40          & 70.10          & 15.01          & 92.26           & 85.00          & 70.50          & 15.23          & 92.98           \\
DoTS \cite{dots}         & 86.59          & 74.14          & 15.06          & 95.43           & 86.65          & 74.18          & 15.90          & 96.32           \\
SOLOIST \cite{soloist}      & 85.50          & 72.90          & 16.54          & 95.74           & -              & -              & -              & -               \\
MinTL \cite{mintl}       & 84.88          & 74.91          & 17.89          & 97.79           & -              & -              & -              & -               \\
UBAR$\dagger$ \cite{ubar}        & 85.10          & 71.02          & 16.21          & 94.27           & 86.20          & 70.32          & 16.48          & 94.74           \\
PPTOD \cite{pptod}       & 89.20          & 79.40          & 18.62          & 102.92          & 87.09          & 79.08          & 19.17          & 102.26          \\
GALAXY (w/o pretrain) \cite{galaxy}       & \textbf{93.10}          & 81.00          & 18.44          & 105.49          & \textbf{93.50}          & 81.70          & 18.32          & 105.92          \\
MTTOD$\ddagger$ \cite{mttod}       & 91.80          & 83.80          & 19.56          & 107.36          & 90.40          & 81.70          & \textbf{20.15} & 106.20          \\
% BORT \cite{bort}        & \textbf{93.80} & \textbf{85.80} & 18.50          & 108.30          & -              & -              & -              & -               \\
\midrule
MDTOD & 92.70          & \textbf{85.00}          & \textbf{19.72} & \textbf{108.57} & 92.70 & \textbf{84.60} & 19.03          & \textbf{107.68} \\
\bottomrule
\end{tabular}
}
\caption{Full dataset comparison results between MDTOD and baselines under end-to-end settings. $\dagger$: the results in \cite{pptod} are utilized. $\ddagger$: reproduced results operating the author's open-source code.}
\label{tab:main_results}
\end{table*}

\section{Case Analysis}
We present partial selections of paraphrases in Table \ref{tab:para_examples} to demonstrate the effect of the rephraser. As shown in the first example, when the constraints are set to the entities "hail" and "los angeles", the rephraser still produces paraphrases that are fluent and satisfy the constraints.

In addition, we illustrate a sample of the dialog generated by MDTOD in Table \ref{tab:case_dialog} . The dialogue begins with the user seeking an Indian restaurant in the center of town, and the model correctly extracts the values of the slots "food" and "area". When the conversation proceeds to turn 2, MDTOD generates more belief states than oracle's belief states, but the model generates the correct results. The reason is that there are some labeling errors in MultiWOZ2.0, while MDTOD can still generate correct belief states, which shows the robustness of MDTOD. When the conversation progressed to turn 5, MDTOD still predicted the correct belief state despite the user changing the reservation time from 13:30 to 12:30, indicating that the model understood the semantic meaning of the current input sentences rather than simply repeating the belief state from the previous turn.

\begin{table*}[t]
\centering
\resizebox{0.90\textwidth}{!}{
\begin{tabular}{ll}
\toprule
                      & \textbf{Examples}                                                \\
\midrule
Constraints           & {[}weather{]} {[}value\_weather\_attribute{]} \textcolor[RGB]{2,195,154}{hail} {[}value\_location{]} \textcolor[RGB]{2,195,154}{los angeles} \\
\midrule
Original Utterance    & is there going to be \textcolor[RGB]{2,195,154}{hail} in \textcolor[RGB]{2,195,154}{los angeles} this weekend ? \\
Original Response     & on Sunday hail is predicted to fall in san mateo        \\
Paraphrased Utterance & will \textcolor[RGB]{2,195,154}{hail} hit \textcolor[RGB]{2,195,154}{los angeles} this weekend?                 \\
Paraphrased Response  & on sunday hail is foreshadow to fall in san mateo       \\
\midrule
Constraints            & {[}schedule{]} {[}value\_event{]} \textcolor[RGB]{2,195,154}{dentist appointment}   \\
\midrule
Original Utterance    & give me the date and time of my \textcolor[RGB]{2,195,154}{dentist appointment}     \\
Original Response    & your dentist appointment is at {[}value\_time{]} on {[}value\_date{]} .              \\
Paraphrased Utterance & tell me the date and time of the \textcolor[RGB]{2,195,154}{dentist appointment}    \\
Paraphrased Response & your tooth doctor appointment is at {[}value\_time{]} on {[}value\_date{]} .         \\
\midrule
Constraints           & {[}schedule{]} {[}value\_party{]} martha {[}value\_event{]} \textcolor[RGB]{2,195,154}{dinner} {[}value\_time{]} \textcolor[RGB]{2,195,154}{11am} \\
\midrule
Original Utterance    & schedule \textcolor[RGB]{2,195,154}{dinner} with Martha later this week at \textcolor[RGB]{2,195,154}{11am}.    \\
Original Response     & what day of the week do you want the reminder set for?  \\
Paraphrased Utterance & plan a \textcolor[RGB]{2,195,154}{dinner} with Martha later this week at \textcolor[RGB]{2,195,154}{11 am}.     \\
Paraphrased Response & what day of the workweek do you want the reminder set for?   \\
\bottomrule
\end{tabular}
}
\caption{A partial sample of the rephrased user utterances and system responses generated by paraphraser, where the constraints are marked in \textcolor[RGB]{2,195,154}{green}.}
\label{tab:para_examples}
\end{table*}

\begin{table*}[t]
\centering
\resizebox{0.95\textwidth}{!}{
\begin{tabular}{cll}
\toprule
\multicolumn{2}{l}{\textbf{SNG0586}} &
  \textbf{Dialogue} \\
\midrule
\multirow{6}{*}{Turn 0} &
  User Utterance &
  hi, i am looking for a restaurant in the centre of town serving north indian food . \\
 &
  Oracle Belief State &
  {[}restaurant{]} \{food : north indian, area : centre\} \\
 &
  Generated Belief State &
  {[}restaurant{]} \{food : north indian, area : centre\} \\
\cmidrule(lr){2-3}
 &
  Oracle Reference &
  we don't have any specific restaurants in that category. let's try something else ? \\
 &
  Delexicalized Response &
  \begin{tabular}[c]{@{}l@{}}i am sorry , there are no restaurants serving {[}value\_food{]} food in the centre .\\  would you like to try a different cuisine ?\end{tabular} \\
 &
  Lexicalized Response &
  \begin{tabular}[c]{@{}l@{}}i am sorry , there are no restaurants serving north indian food in the centre .\\  would you like to try a different cuisine ?\end{tabular} \\
\midrule
\multirow{6}{*}{Turn 1} &
  User Utterance &
  that 's too bad . how about chinese ? \\
 &
  Oracle Belief State &
  {[}restaurant{]} \{food : chinese, area : centre\} \\
 &
  Generated Belief State &
  {[}restaurant{]} \{food : chinese, area : centre\} \\
\cmidrule(lr){2-3}
 &
  Oracle Reference &
  okay sure what do you have ? \\
 &
  Delexicalized Response &
  \begin{tabular}[c]{@{}l@{}}there are {[}value\_choice{]} {[}value\_food{]} restaurants in the {[}value\_area{]} . do you\\  have a price range preference ?\end{tabular} \\
 &
  Lexicalized Response &
  there are 3 chinese restaurants in the centre . do you have a price range preference ? \\
\midrule
\multirow{6}{*}{Turn 2} &
  User Utterance &
  \begin{tabular}[c]{@{}l@{}}okay when you find something i would like to make a reservation for 2 people\\  on a sunday at 13:30 .\end{tabular} \\
 &
  Oracle Belief State &
  {[}restaurant{]} \{food : chinese, area : centre\} \\
 &
  Generated Belief State &
  {[}restaurant{]} \{food : chinese, area : centre, \textcolor[RGB]{2,195,154}{time : 13:30, day : sunday, people : 2}\} \\
  \cmidrule(lr){2-3}
 &
  Oracle Reference &
  do you have a price range in mind ? \\
 &
  Delexicalized Response &
  \begin{tabular}[c]{@{}l@{}}i was able to book you at {[}value\_name{]} . your reference number is \\ {[}value\_reference{]} . is there anything else i can help you with ?\end{tabular} \\
 &
  Lexicalized Response &
  \begin{tabular}[c]{@{}l@{}}i was able to book you at charlie chan . your reference number is 4wngilmf .\\  is there anything else i can help you with ?\end{tabular} \\
\midrule
\multirow{6}{*}{Turn 3} &
  User Utterance &
  i don't mind about the price range . \\
 &
  Oracle Belief State &
  {[}restaurant{]} \{food : chinese, area : centre\} \\
 &
  Generated Belief State &
  {[}restaurant{]} \{food : chinese, area : centre, \textcolor[RGB]{2,195,154}{time : 13:30, day : sunday, people : 2}\} \\
  \cmidrule(lr){2-3}
 &
  Oracle Reference &
  \begin{tabular}[c]{@{}l@{}}okay , how about lan hong house ? it is a chinese restsurant in the centre area\\  and is in the moderate price range . will that work for you ?\end{tabular} \\
 &
  Delexicalized Response &
  i would recommend {[}value\_name{]} . \\
 &
  Lexicalized Response &
  i would recommend lan hong house . \\
\midrule
\multirow{6}{*}{Turn 4} &
  User Utterance &
  that 's fine . can you try booking it ? \\
 &
  Oracle Belief State &
  {[}restaurant{]} \{food : chinese, area : centre, time : 13:30, day : sunday, people : 2\} \\
 &
  Generated Belief State &
  {[}restaurant{]} \{food : chinese, area : centre, time : 13:30, day : sunday, people : 2\} \\
  \cmidrule(lr){2-3}
 &
  Oracle Reference &
  \begin{tabular}[c]{@{}l@{}}i am sorry , but i was n't able to book the restaurant for that date and time .\\  would you like to try a different time ?\end{tabular} \\
 &
  Delexicalized Response &
  \begin{tabular}[c]{@{}l@{}}booking was successful . the table will be reserved for 15 minutes . reference\\  number is : {[}value\_reference{]} . anything else i can help with ?\end{tabular} \\
 &
  Lexicalized Response &
  \begin{tabular}[c]{@{}l@{}}booking was successful . the table will be reserved for 15 minutes . reference\\  number is : f1ntkwzo . anything else i can help with ?\end{tabular} \\
\midrule
\multirow{6}{*}{Turn 5} &
  User Utterance &
  how about 12:30 ? \\
 &
  Oracle Belief State &
  \begin{tabular}[c]{@{}l@{}}{[}restaurant{]} \{food : chinese, area : centre, name : lan hong house, time : 12:30,\\  day : sunday, people : 2\}\end{tabular} \\
 &
  Generated Belief State &
  {[}restaurant{]} \{food : chinese, area : centre, time : 12:30, day : sunday, people : 2\} \\
  \cmidrule(lr){2-3}
 &
  Oracle Reference &
  \begin{tabular}[c]{@{}l@{}}you are booked for 12:30 on sunday for 2 people . your reference number is \\ f1ntkwzo . is there something else i can help you with ?\end{tabular} \\
 &
  Delexicalized Response &
  \begin{tabular}[c]{@{}l@{}}booking was successful . the table will be reserved for 15 minutes . reference\\  number is : {[}value\_reference{]} . anything else i can help with ?\end{tabular} \\
 &
  Lexicalized Response &
  \begin{tabular}[c]{@{}l@{}}booking was successful . the table will be reserved for 15 minutes . reference\\  number is : f1ntkwzo . anything else i can help with ?\end{tabular} \\
\midrule
\multirow{6}{*}{Turn 6} &
  User Utterance &
  thanks so much . have a nice day . bye . \\
 &
  Oracle Belief State &
  \begin{tabular}[c]{@{}l@{}}{[}restaurant{]} \{food : chinese, area : centre, name : lan hong house, time : 12:30,\\  day : sunday, people : 2\}\end{tabular} \\
 &
  Generated Belief State &
  {[}restaurant{]} \{food : chinese, area : centre, time : 12:30, day : sunday, people : 2\} \\
  \cmidrule(lr){2-3}
 &
  Oracle Reference &
  thank you , if you need anything else let me know . good bye . \\
 &
  Delexicalized Response &
  thank you for using our system ! \\
 &
  Lexicalized Response &
  thank you for using our system !  \\
\bottomrule
\end{tabular}
}
\caption{A dialogue sample generated by MDTOD on MultiWOZ 2.0.}
\label{tab:case_dialog}
\end{table*}

\end{document}